\documentclass[lettersize,journal]{IEEEtran}
\usepackage{amsmath,amsfonts}
\usepackage{algorithmic}
\usepackage{algorithm}
\usepackage{array}
\usepackage[caption=false,font=normalsize,labelfont=sf,textfont=sf]{subfig}
\usepackage{textcomp}
\usepackage{stfloats}
\usepackage{url}
\usepackage{verbatim}
\usepackage{graphicx}
\usepackage{cite}
\usepackage{graphicx}
\usepackage{amsmath}
\usepackage{amssymb}
\usepackage{booktabs}
\usepackage{cleveref}
\usepackage{bm}
\usepackage{multirow}
\usepackage{shadow}
\usepackage{bbding}
\usepackage{pifont}
\usepackage{colortbl}
\usepackage{pifont}
\definecolor{graymy}{RGB}{200,200,200}
\begin{document}

\title{Attention-disentangled Uniform Orthogonal Feature Space Optimization for Few-shot Object Detection}

\author{Taijin Zhao, Heqian Qiu, \textit{Member, IEEE}, Yu Dai, Lanxiao Wang, Fanman Meng, \textit{Member, IEEE}, Qingbo Wu, \textit{Member, IEEE} and Hongliang Li, \textit{Senior Member, IEEE}
\thanks{Taijin Zhao, Heqian Qiu, Yu Dai, Lanxiao Wang, Fanman Meng, Qingbo Wu and Hongliang Li are with the University of Electronic Science and Technology of China, Chengdu, 611731, China (emall: zhtjww@std.uestc.edu.cn, hqqiu@uestc.edu.cn, ydai@std.uestc.edu.cn, lanxiaowang@uestc.edu.cn, fmmeng@uestc.edu.cn, qbwu@uestc.edu.cn, hlli@uestc.edu.cn)}}

\markboth{Journal of \LaTeX\ Class Files}%
{Shell \MakeLowercase{\textit{et al.}}: A Sample Article Using IEEEtran.cls for IEEE Journals}


\maketitle

\begin{abstract}
Few-shot object detection (FSOD) aims to detect objects with limited samples for novel classes, while relying on abundant data for base classes. Existing FSOD approaches, predominantly built on the Faster R-CNN detector, entangle objectness recognition and foreground classification within shared feature spaces. This paradigm inherently establishes class-specific objectness criteria and suffers from unrepresentative novel class samples. To resolve this limitation, we propose a Uniform Orthogonal Feature Space (UOFS) optimization framework. First, UOFS decouples the feature space into two orthogonal components, where magnitude encodes objectness and angle encodes classification. This decoupling enables transferring class-agnostic objectness knowledge from base classes to novel classes. Moreover, implementing the disentanglement requires careful attention to two challenges: (1) Base set images contain unlabeled foreground instances, causing confusion between potential novel class instances and backgrounds. (2) Angular optimization depends exclusively on base class foreground instances, inducing overfitting of angular distributions to base classes. To address these challenges, we propose a Hybrid Background Optimization (HBO) strategy: (1) Constructing a pure background base set by removing unlabeled instances in original images to provide unbiased magnitude-based objectness supervision. (2) Incorporating unlabeled foreground instances in the original base set into angular optimization to enhance distribution uniformity. Additionally, we propose a Spatial-wise Attention Disentanglement and Association (SADA) module to address task conflicts between class-agnostic and class-specific tasks. Experiments demonstrate that our method significantly outperforms existing approaches based on entangled feature spaces.

\end{abstract}

\begin{IEEEkeywords}
Few-shot object detection, Object detection, Feature orthogonality, Task disentanglement.
\end{IEEEkeywords}

\section{Introduction}
\label{sec:intro}
Object detection is a fundamental computer vision task that localizes and identifies objects of predefined classes within images. Unlike conventional object detection methods \cite{redmon2017yolo9000,qiu2022crossdet++,girshick2015fast,ren2016faster,cai2018cascade} requiring extensive training samples per class, few-shot object detection (FSOD) \cite{wang2020frustratingly,qiao2021defrcn,cao2021fadi,kaul2022label,tang2023semi,zhao2024vlm,xin2024ecea} trains detectors with a limited number of samples, while also leveraging abundant auxiliary samples from other classes. Classes with only a few samples are termed novel classes, while classes with abundant samples are termed base classes. FSOD plays a crucial role in diverse data scarce scenarios, such as industrial applications, transportation systems, and rare species conservation.

\begin{figure}[!t]
    \centering
     \includegraphics[width=1.0\columnwidth]{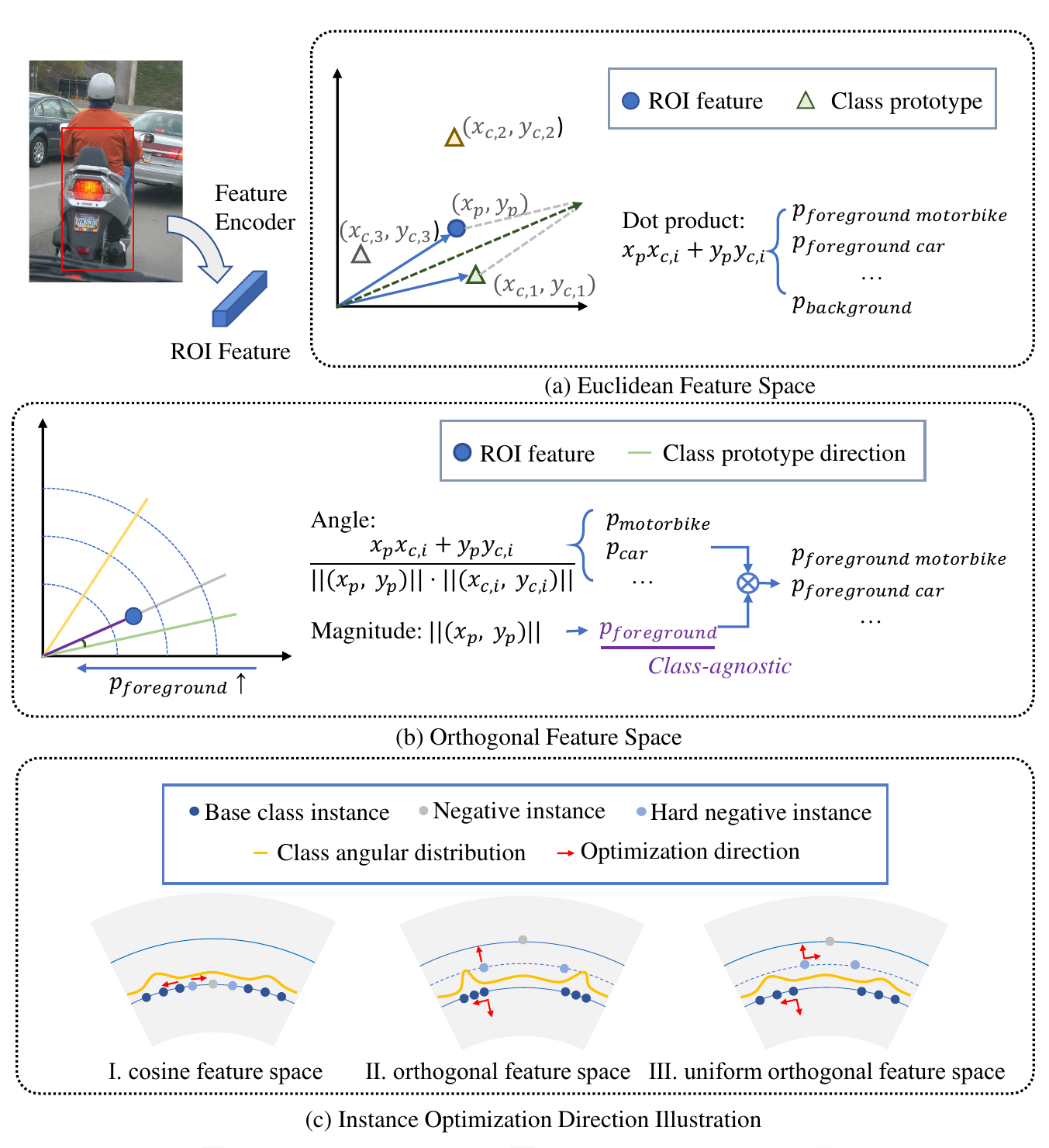}
    \caption{(a) In entangled feature space (e.g., Euclidean space), where background is treated as a special foreground class, each foreground class prediction reflects the objectness and foreground class probabilities simultaneously. (b) In the orthogonal feature space, objectness and foreground class probability predictions are decoupled: magnitude encodes objectness while angle encodes classification. This disentanglement establishes class-agnostic objectness criteria shared across all foreground classes, thereby allowing effective transfer of objectness recognition capabilities from base to novel classes. (c) By involving negative samples in angular optimization, uniform orthogonal feature space utilizes the class information within hard negative samples (unlabeled samples with semantics similar to potential novel classes) to mitigate the angular distribution overfitting to base classes.}
    \label{fig: intro1}
\end{figure}

FSOD methods are typically built upon conventional object detection frameworks with adaptations to accommodate few-shot conditions. Recent advanced FSOD approaches primarily adopt the Faster R-CNN \cite{ren2016faster} detector due to its powerful detection capabilities. Faster R-CNN detector first generates multiple region proposals and extracts their ROI features, which are then processed through Fully Connected (FC) or cosine classifiers to compute classification scores for foreground classes and background. FC and cosine classifiers construct Euclidean and cosine feature spaces, respectively. The classification score represents the confidence that a proposal belongs to a ``foreground [CLS]'', comprising two recognition properties: (1) objectness: the confidence score of a proposal being a foreground instance, and (2) class probability: the classification probabilities across all foreground classes without considering objectness. While classification relies on high-level semantics, objectness recognition mainly depends on low-level features (e.g., texture, color) without considering the class of proposals. However, in these methods, the recognition criteria of objectness are class-specific, and are trained with few novel class samples for each class in isolation. These few novel class instances tend to exhibit individuality, consequently failing to capture the complete class distribution. In Fig. \ref{fig: intro1}, for example, the motorbike suffers from incompleteness due to the block of other instances. As shown in Fig. \ref{fig: intro1} (a), taking Euclidean space as an example, by treating background as a special case of foreground classes, objectness and classification become entangled in a shared criterion, and the detector directly recognizes whether a proposal is a ``foreground motorbike'', a ``foreground car'' etc. Fine-tuning with these biased samples causes the detector to misinterpret partial instances as complete instances, resulting in mislocalization during testing.

In this paper, we address this problem from the perspective of recognition criteria and feature spaces, proposing a Uniform Orthogonal Feature Space (UOFS). Inspired by orthogonal feature space (OFS) \cite{sun2024exploring,chen2020norm}, we first disentangle objectness and foreground classification prediction via orthogonal components in the feature space for FSOD. As shown in Fig. \ref{fig: intro1} (b), using feature's magnitude to predict objectness and using the angle between feature and class prototypes to predict the class label. Unlike entangled feature spaces (EFS), which establish class-specific objectness criteria for each category, OFS separates objectness measurement from classification, which enables the detector to establish a magnitude-based objectness criterion shared across all foreground classes. Consequently, objectness recognition capability learned from base classes can be effectively transferred to novel classes, thereby mitigating objectness recognition bias arising from unrepresentative novel class training samples. 

While theoretically promising, the simple disentanglement faces practical limitations in the common FSOD training pipeline \cite{wang2020frustratingly}, where the detector is pretrained with abundant base class samples and fine-tuned with few-shot novel class samples. To avoid overfitting to scarce novel class samples, parameters of the feature extractor are largely frozen during fine-tuning. Consequently, the feature space is primarily determined by the base training. Two critical limitations emerge when combining this paradigm with OFS. The first is label missing: Foreground instances excluded from base classes are unlabeled and regarded as backgrounds, and these instances may later become novel classes or exhibit high semantic similarities to potential novel classes. Optimizing their magnitudes to diverge from foreground instances suppresses their feature expressiveness, leading to misclassification of semantically similar novel class instances as backgrounds. The second is angular overfitting: As shown in Fig. \ref{fig: intro1} (c), entangled feature spaces (e.g., cosine space) treat background as a special foreground class, forcing unlabeled hard negatives (semantically similar to base classes) away from foreground class clusters. In contrast, OFS constrains negative sample updates exclusively to the magnitude dimension without affecting angular distributions, discarding valuable semantic information in hard negatives. Consequently, unlabeled samples converge toward base class prototypes due to missing adversarial calibration, making potential novel class instances collapse into base class clusters, ultimately degrades novel class generalization.

To address the aforementioned challenges, we propose a Hybrid Background Optimization (HBO) framework that enables unlabeled foreground instances in base set images to contribute to angular optimization without corrupting magnitude optimization. First, we generate a pure background base set by cropping and pasting annotated base class instances from base set images onto gray backgrounds, eliminating unlabeled instances in the original base set images. Negative proposals in the pure background base set are used to optimize the magnitude as they don't contain any unlabeled instances disturbing the objectness recognition criteria. Meanwhile, negative proposals from the original base set are used to optimize both magnitude and angle. With parameter sharing, magnitude consistency between the two sets identifies candidate proposals with potential unlabeled foreground instances, prioritizing them for angular optimization. By optimizing magnitude and angle via distinct negative sampling strategies, HBO ensures robust objectness recognition and achieves angular geometric stability.

Moreover, we analyze the task conflicts and associations among objectness recognition, classification, and box regression. While objectness recognition and box regression rely on class-agnostic semantic representations, classification necessitates class-specific patterns. To resolve this discrepancy, we propose a Spatial-wise Attention Disentanglement and Association (SADA) module to generate task-specific features tailored for class-agnostic and class-specific tasks through spatial attention mechanisms.

This paper makes the following contributions:
\begin{itemize}
    \item We systematically analyze the vulnerability of entangled feature spaces caused by unrepresentative novel class samples in FSOD.
    \item We propose a uniform orthogonal feature space with hybrid background optimization to realize the objectness transfer and improve classification generalization to novel classes.
    \item We design a spatial-wise attention disentanglement and association module to decouple and associate representations for class-agnostic and class-specific tasks.
\end{itemize}

\section{Related Works}
\subsection{Object Detection}
Conventional deep learning based object detection methods focus on localizing foreground objects and identifying their categories within images, assuming the availability of abundant training samples for each class. These approaches are categorized into two main types: one-stage detectors \cite{redmon2017yolo9000,qiu2022crossdet++} and two-stage detectors \cite{girshick2015fast,ren2016faster,cai2018cascade}. FSOD methods typically extend conventional detectors by adapting them to operate in scenarios with limited training data. A few early FSOD approaches \cite{kang2019few, li2021beyond} employ YOLO as their base architecture, which uses separate heads to predict objectness and foreground class probabilities. In recent years, owing to its superior performance, Faster R-CNN \cite{ren2016faster} has become the predominant basic detector for FSOD frameworks. While leveraging Faster R-CNN’s robust detection capabilities, recent FSOD methods sacrifice YOLO’s ability to predict class-agnostic objectness. Furthermore, YOLO’s disentanglement mechanism operates solely at the prediction level and lacks explicit orthogonality constraints in the feature space, thus failing to eliminate inherent correlations between objectness and foreground class probabilities.

\subsection{Few-shot Object Detection}
Few-shot object detection (FSOD) extends conventional object detection to sample scarcity scenarios, where base class samples are abundant while novel class samples are scarce. Researchers develop episode task methods and single task methods for FSOD \cite{xin2024few}. Both paradigms address novel class data scarcity through a shared principle: transferring knowledge from base to novel classes. Episode task methods employ a meta-learning framework, partitioning base class samples into episodic groups. Through iterative training on these episodes, the detector acquires class-agnostic ``meta-knowledge'', enabling rapid adaptation to novel classes with limited samples. 
In contrast, single task methods bypass episodic training. TFA \cite{wang2020frustratingly} first demonstrated that pretraining a detector on base class samples and fine-tuning it with few-shot novel class samples achieves strong performance on novel classes. Single task methods subsequently became the dominant paradigm. Building on this pipeline, following studies explored enhancing knowledge transfer mechanisms from base classes to novel classes. DeFRCN \cite{qiao2021defrcn} proposes a gradient decouple layer to scale the learning rate of different components of the Faster R-CNN detector, preserving base class knowledge while retaining plasticity for novel class fine-tuning. FADI \cite{cao2021fadi} links each novel class to the most similar base class to imitate the well learned feature space of base classes. Semi-supervised methods \cite{kaul2022label,tang2023semi,zhao2024vlm} assume that base images contain unlabeled novel class instances and employ pseudo-labeling strategies to transfer explicit novel class semantics from base data to the detector. ECEA \cite{xin2024ecea} addresses the novel class sample incompleteness via an extensible coexisting attention module to associate instance parts and whole instances, which still captures class-specific semantics for objectness recognition. Class-agnostic objectness migration is hardly mentioned in previous FSOD methods, which largely contributes to distinguishing novel class instances from backgrounds.

\subsection{Orthogonal Feature Space}
The geometry of the feature space is critical to representation learning, with orthogonal feature space serving as a prominent framework for disentangling latent components. Researches on orthogonality in few-shot learning primarily focus on creating class-wise \cite{simon2020adaptive,hersche2022constrained,ranasinghe2021orthogonal} or base-novel-class \cite{liu2023learning} subspaces to enhance inter-class discriminability and resolve base-novel class conflicts. Norm-VAE \cite{xu2023normvae} constructs an orthogonal feature space for the latent codes of the variational autoencoder but neglects orthogonal constraints in the final decision space, leading to a mismatch between latent and final feature distributions. Our work draws inspiration from OrthogonalDet \cite{sun2024exploring}, which leverages orthogonal feature spaces to decouple objectness and classification predictions. While OrthogonalDet emphasizes class-agnostic components to improve transferability, we extend this concept to FSOD to mitigate novel class sample unrepresentativeness. Moreover, compared to the strategy of selecting top-K objectness proposals as unseen foreground objects \cite{sun2024exploring}, our hybrid background optimization framework generates more uniform angular feature distributions, improving novel class generalization.

\subsection{Crop-Paste Augmentation}
The crop-paste operation involves extracting foreground instances via bounding box or mask annotations and pasting them onto another image. This technique is a widely utilized data augmentation strategy in instance segmentation \cite{ghiasi2021simple,zhao2023x}, object localisation \cite{dorkenwald2024pin}, and object detection \cite{lin2023effective,saito2022learning}. The most relevant prior work to our approach is LDET \cite{saito2022learning}, which employs crop-paste augmentation to erase unannotated instances for open-world object detection. Compared to LDET, our method incorporates orthogonal feature space learning to explicitly transfer objectness recognition capabilities. Furthermore, while LDET overlooks unannotated instances that can enhance feature space uniformity, our framework leverages these instances to construct more uniform angular distributions, resulting in enhanced generalization to novel classes.

\section{Preliminary}
\subsection{Task Setting}
The task configuration follows prior few-shot object detection studies \cite{wang2020frustratingly,qiao2021defrcn,cao2021fadi,kaul2022label,tang2023semi,zhao2024vlm,xin2024ecea}. The dataset consists of two mutually exclusive subsets: the base class subset, denoted as $D_B$, and the novel class subset, denoted as $D_N$. The base class subset $D_B$ encompasses a copious quantity of instances for each class, whereas the novel class set $D_N$ is composed of a limited number of \textit{k} samples per class within the \textit{k}-shot framework. We adopt the G-FSOD setting, where foreground classes include all base and novel classes in the final detector. 
 
\subsection{Entangled Feature Space}
Faster R-CNN based method TFA \cite{wang2020frustratingly} utilizes either fully connected or cosine classifiers to construct entangled feature spaces for classification. This strategy is also adopted in subsequent works \cite{sun2021fsce,qiao2021defrcn,xu2023normvae}. The classification score for each proposal can be computed by:
\begin{equation}
S_{p, i}=\mathrm{Softmax}\left(d\left(\bm{W}_i, \bm{f}\right)\right)
\end{equation}
where $\bm{f}$ is the ROI feature of a proposal, $\bm{W}_i$ is the learnable class prototype corresponding to i-th class (including all foreground classes and the background class). $d\left(\cdot\right)$ denotes the distance metric, which is the dot product for the FC classifier and the cosine similarity for the cosine classifier, respectively.

$S_{p, i}$ simultaneously encodes the objectness and classification confidence. Since the classification scoring of each category is independent, the objectness of each class is also measured independently. When novel class samples are unrepresentative of their class distributions, the classifier relies on biased instance features to establish criteria for calculating the objectness, leading to the classifier's inability to determine the completeness of the novel class instances.

\begin{figure*}[!t]
    \centering
     \includegraphics[width=2.0\columnwidth]{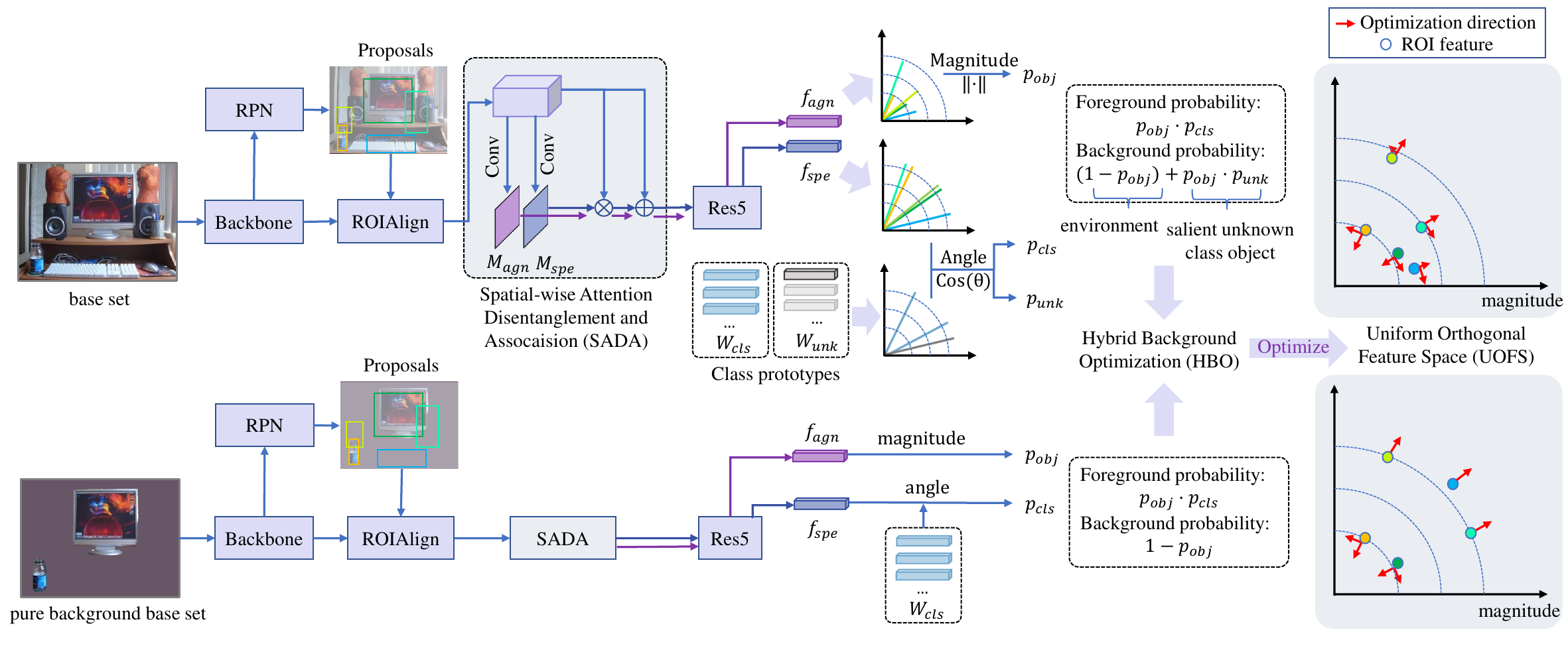}
    \caption{The framework of our proposed method. Images from the base set and the pure background base set are sent into the detector simultaneously. The pure background base set is adopted to learn the magnitude-based objectness recognition ability while negative samples in the base set are extra adopted to optimize the angular distribution. Moreover, a spatial-wise attention disentanglement and association module is proposed to extract class-agnostic and class-specific features for subsequent tasks.}
    \label{fig: main}
\end{figure*}

\section{Method}

\subsection{Overview}
The framework of our proposed method is shown in Fig. \ref{fig: main}. Following previous works \cite{wang2020frustratingly,sun2021fsce,qiao2021defrcn,wang2025orthogonal,xin2024ecea}, our method is built upon the Faster R-CNN detector with the base-novel training paradigm. In this section, we first present how to disentangle objectness and classification in Section \ref{method:decouple}. Section \ref{method:pbbs} introduces a pure background base set to eliminate unlabeled objects in base images. Section \ref{method:hbo} details training the detector with a hybrid background optimization framework, combining the pure background set and original base set. Moreover, we discuss the task conflict of ROI features and propose a spatial-wise attention disentanglement and association module in Section \ref{method:sada}. Finally, we provide a numerical comparison between orthogonal and entangled feature spaces for few-shot object detection in Section \ref{method:compare}.

\subsection{Disentangle Objectness and Classification}
\label{method:decouple}
Unlike approaches that are jointly modeling objectness and classification, some previous works disentangle the computation of objectness and classification \cite{ma2023cat,sun2024exploring}, and further combine with orthogonal feature space \cite{sun2024exploring}, using the magnitude to decide the objectness while using the angle to decide the class. The objectness probability $p_{\text{obj}}$ and the foreground class probabilities $p_{\text{cls}}$ are computed as:
\begin{equation}
    p_{\text{obj}} = h_{\text{obj}}\left(\|\bm{f}\|\right), p_{\text{cls}} = h_{\text{cls}}\left(\frac{\bm{f}}{\|\bm{f}\|}\right)
\end{equation}

Then the joint class probabilities for each proposal can be computed by: \begin{equation}
p_{\text{joint}}=p_{\text{obj}}\cdot p_{\text{cls}}
\end{equation}

In the disentangled formulation, objectness is class-agnostic, and can be optimized by all class samples, enabling effective transfer of foreground-background discrimination ability from base to novel classes.

\subsection{Pure Background Base-set Construction}
\label{method:pbbs}
The primary challenge in base training stems from unlabeled instances and real backgrounds being conflated as ``background''. As novel class instances may share semantics with unlabeled instances, regarding them as background will fall to recall potential novel class instances. Inspired by \cite{dorkenwald2024pin,saito2022learning}, we propose to construct a pure background base set (PB base set) to assist the learning of objectness criterion. 

\begin{figure}[!t]
    \centering
     \includegraphics[width=0.8\columnwidth]{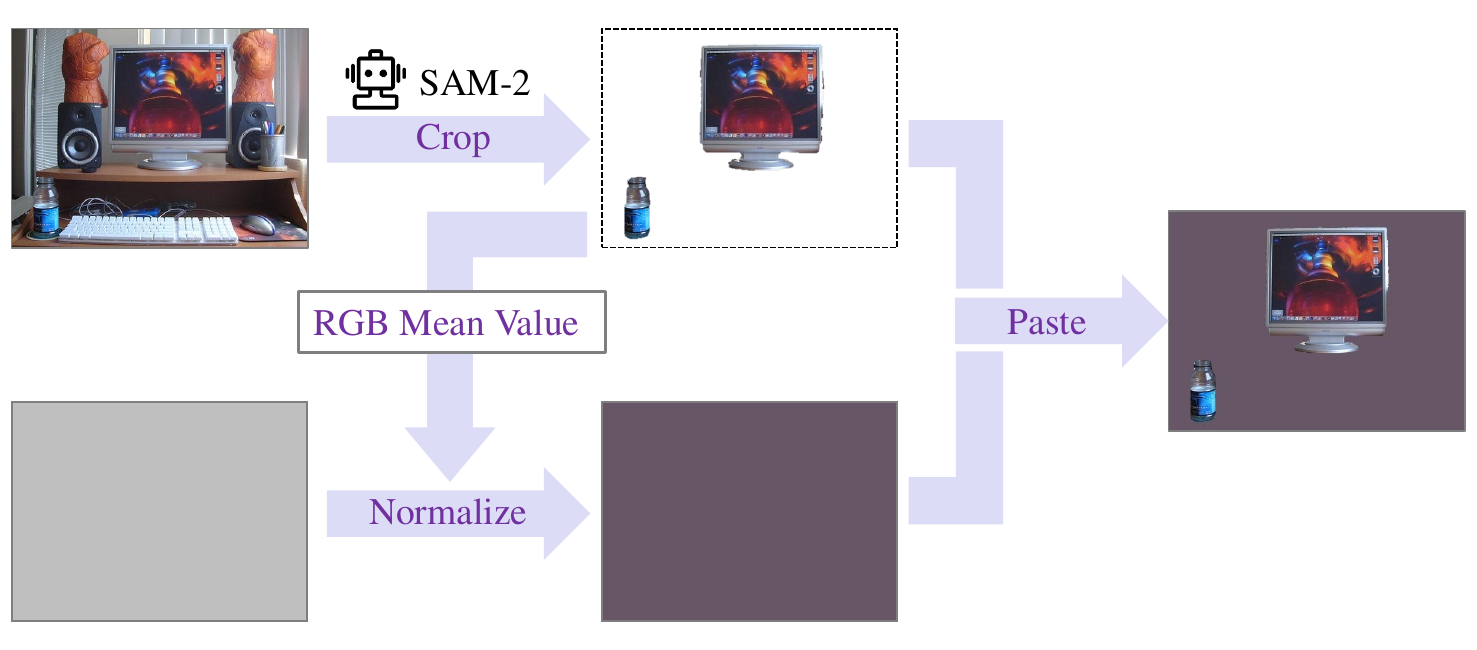}
    \caption{The pipeline of constructing the pure background base set.}
    \label{fig: pbbs}
\end{figure}

As shown in Fig. \ref{fig: pbbs}, following the crop-paste pipeline \cite{dorkenwald2024pin, lin2023effective}, the process involves two steps. The first is separating out foreground base class instances from base set images, and the second is pasting foreground instances onto background images that do not contain salient foreground instances. For a base image, we use a zero-shot segmentation model SAM-2 \cite{ravi2024sam2} to get base class instance masks. The model takes base set images and annotated base class bounding boxes as prompts to precisely segment foreground objects. The foreground instances are then pasted onto gray background images with the same height and width with the original images. We design three background types: (1) fixed gray backgrounds with RGB value [127,127,127], (2) normalized gray backgrounds that have the same mean value with the foreground instances, (3) resized background images randomly selected from the BG-20k dataset \cite{li2022bridging}. We find that normalized gray backgrounds achieve the best performance. Compared to (1), this result demonstrates the importance of introducing variations in the background. Compared to (3), natural scene background images from BG-20k still contain some foreground instances that obstruct objectness optimization. Further analysis of the experimental results can be found in Section \ref{sec:exp-ablation}.

\subsection{Hybrid Background Optimization}
\label{method:hbo}
In base set images, background proposals can be classified into two types: (1) true background (environment regions without salient objects), and (2) unlabeled instances (salient instances excluded from base class annotations). In the orthogonal feature space, true backgrounds should be separated from foregrounds via magnitude differences, while unlabeled instances should maintain magnitude similarity but differ in angular distribution. Directly optimizing unlabeled instances to differentiate their feature magnitudes from base class foreground instances would restrict the detector's capacity to generalize beyond base classes.

A straightforward operation is to directly train the detector solely with the constructed PB base set. However, this faces three challenges: Firstly, there are domain gaps due to the different background sources of base set and PB base set. Secondly, SAM-2's imperfect segmentation yields incomplete foreground objects. Thirdly, absence of hard negatives in angular optimization causes overfitting.

In this paper, we propose a joint training strategy. In a mini-batch, an image from base set and another image from PB base set are sent into the detector. Losses for the two images are computed separately, losses for the base set and the PB base set are denoted as $L_{\text{det-BS}}$ and $L_{\text{det-PBBS}}$. Then the two losses are integrated to generate the final loss:
\begin{equation}
    L_{\text{det}} = \alpha L_{\text{det-BS}} + (1-\alpha) L_{\text{det-PBBS}}
\end{equation}
where $\alpha$ is a hyperparameter that controls the importance of base set and PB base set, empirically set to 0.5 in this paper.

The objectness score $p_{\text{obj}}$ in $\mathbb{R}^{1\times1}$ is computed as: 
\begin{equation}
\begin{aligned}
p_{\text{obj}} &= h_{\text{obj}}(\| \bm{f} \|) \\
&= -|\bm{W}_{\text{obj}}| \cdot \| \bm{f} \| + b
\end{aligned}
\end{equation}
where $\bm{W}_{\text{obj}}$ and $b$ are learnable parameters in $\mathbb{R}^{1\times1}$, $|\cdot|$ represents computing absolute value, the operation ensures that instances with smaller magnitudes have larger objectness scores, the benefits are further discussed in Section \ref{sec:exp-ablation}.

For ROI feature $\bm{f}$ in $\mathbb{R}^{1\times 2048}$, the cosine similarity based classification probabilities $p_{\text{cls}}$ in $\mathbb{R}^{1\times N_c} $ for all $N_c$ foreground classes can be computed as:
\begin{equation}
\begin{aligned}
p_{\text{cls}} &= h_{\text{cls}} \left( \frac{\bm{f}}{\| \bm{f} \|} \right) \\
&= \mathrm{Softmax} \left( \frac{\tau \bm{f} \bm{W}_{\text{cls}}}{\| \bm{f} \| \| \bm{W}_{\text{cls}} \|} \right)
\end{aligned}
\end{equation}
where $W_{\text{cls}}$ in $\mathbb{R}^{2048\times N_c}$ are learnable base class prototypes, $\tau$ is the temperature, empirically set to 20 following \cite{wang2020frustratingly}.

The joint objectness-aware class probabilities $p_{\text{joint}}$ in $\mathbb{R}^{1\times\left(N_c+1\right)}$ for all foreground class and the background class can be integrated by:
\begin{equation}
\label{equ: joint1}
    p_{\text{joint}} = \left[p_{\text{cls}} \cdot p_{\text{obj}},1-p_{\text{obj}}\right]
\end{equation}
where $[\cdot]$ denotes concat operation. Then we can compute the classification loss for PB base set by the cross entropy loss between $p_{\text{joint}}$ and the ground truth classification labels. The optimization goal for PB base set is to minimize foreground ROI features' magnitudes and align with corresponding class prototypes in angle, and maximize background ROI features' magnitudes without angular constraints.

For the base set, to represent two types of backgrounds, we propose a hybrid background representation method. We first define a set of learnable prototypes $\bm{W}_{\text{unk}}$ in $\mathbb{R}^{2048 \times N_u}$ to represent $N_u$ unknown classes. For each foreground proposal in the base set, we select a most similar unknown class by sorting the cosine similarity between $\bm{f}$ and $\bm{W}_{\text{unk}}$, the selected unknown class prototype is denoted as $\bm{W}_{\text{unk}}^{\text{m}}$ in $\mathbb{R}^{2048 \times 1}$. The classification probabilities $p_{\text{cls-unk}}$ in $\mathbb{R}^{1\times(N_c+1)}$ over all foreground classes and the most probable unknown class can be computed by:
\begin{equation}
\begin{aligned}
    p_{\text{cls-unk}} &= h_{\text{cls-unk}}\left(\frac{\bm{f}}{\|\bm{f}\|}\right) \\
    &= \mathrm{Softmax}\left(\frac{\tau \bm{f} [\bm{W}_{\text{cls}},\bm{W}_{\text{unk}}^{\text{m}}]}{\|\bm{f}\| \|[\bm{W}_{\text{cls}},\bm{W}_{\text{unk}}^{\text{m}}]\|}\right)
\end{aligned}
\end{equation}

The first $N_c$ terms of $p_{\text{cls-unk}}$ are denoted as $p_{\text{cls-unk}}[:,:N_c]$, which represents the foreground class probabilities, and the last term of $p_{\text{cls-unk}}$ is denoted as $p_{\text{cls-unk}}[:,-1]$, which represents the unknown class probability. During base training, unknown class instances are also labeled as backgrounds, and the joint objectness-aware class probabilities can be integrated by:
\begin{equation}
\begin{aligned}
    p_{\text{joint}} = \left[\, p_{\text{cls}}[:,:N_c] \cdot p_{\text{obj}},\ 1-p_{\text{obj}} + p_{\text{cls-unk}}[:,-1] \cdot p_{\text{obj}} \,\right]
\end{aligned}
\end{equation}

The benefit of the method is that negative samples are adopted to adversarial to positive samples in magnitude and angle simultaneously. For background environments without salient objects, giving them large $p_{\text{cls-unk}}[:,-1]$ and small $p_{\text{obj}}$ will conflict with the optimizing goal of PB base set training, thus the objectness learning automatically aligns with PB base set. Similarly, for salient objects excluded from base classes, predicting large $p_{\text{obj}}$ will conflict with PB base set, as objectness information is usually shared among seen and unseen classes. To align with PB base set training, these hidden objects are forced to have large $p_{\text{cls-unk}}[:,-1]$, achieving more uniform feature distribution in angle. Moreover, for dual attribute negative samples with intermediate value of $p_{\text{obj}}$, they are still forced to have proper $p_{\text{cls-unk}}[:,-1]$ to optimize the angle of positive and negative samples.

During base training, HBO is adopted to improve the feature space generalization. During novel training, we directly utilize the novel set to fine-tune the detector with the joint classification probability formulated in Equ. \ref{equ: joint1}, as novel class instances are already annotated in the novel set.

\subsection{Spatial-wise Attention Disentanglement and Association}
\label{method:sada}
In the Faster R-CNN detector, ROI features are subsequently adopted for classification and regression. Some previous studies \cite{wu2022mfdc,han2023vfa,wang2025orthogonal} point out that the two tasks are conflicting and propose to generate task-specific representations in different ways. In UOFS, original objectness-aware classification is disentangled into objectness-agnostic classification and objectness recognition. To better discriminate the categories of proposals, ROI features should capture class-specific semantics in proposals. However, for objectness recognition and box regression, focusing solely on discriminative semantics will fail to determine the completeness of objects. In contrast, objectness recognition and regression should pay more attention to global structure information. Inspired by attention mechanisms for object detection \cite{zhao2024vlm,qiu2024mcce}, we propose a Spatial-wise Attention Disentanglement and Association (SADA) module  to extract task-specific and task-agnostic representations for subsequent tasks. Given the ROI feature $\bm{F}_4 \in \mathbb{R}^{1024\times7\times7}$ before the Res5 layer, the attention masks $M_\mathrm{spe}, M_\mathrm{agn}$ are computed by:
\begin{align}
    \bm{M}_{\text{spe}} &= \mathrm{Sigmoid}(\bm{W}_{\text{spe}}\bm{F}_4+b_{\text{spe}}) \\
    \bm{M}_{\text{agn}} &= \mathrm{Sigmoid}(\bm{W}_{\text{agn}}\bm{F}_4+b_{\text{agn}})
\end{align}
where $\bm{W}_{\{\text{spe,agn}\}}$ in $\mathbb{R}^{1\times1024}$ ,$b_{\{\text{spe,agn}\}}$ in $\mathbb{R}^{1\times1}$ are the weights and bias of $1\times1$ convolution layers. Then we can compute the enhanced class-specific ROI feature $\bm{f}_\mathrm{spe}$ and class-agnostic ROI feature $\bm{f}_{\text{agn}}$ by multiplying $\bm{F}_4$ with $\bm{M}_\mathrm{\{spe,agn\}}$ in a residual manner and then send the attended feature into Res5 layer. $\bm{f}_{spe}$ is adopted for objectness-agnostic classification while $\bm{f}_{agn}$ is adopted for objectness recognition and regression for associating the two class-agnostic tasks through attention sharing.

\subsection{Numerical Comparison between OFS and EFS for FSOD}
\label{method:compare}
Given ROI feature $\bm{f}$ and class prototype $\bm{W}_i$ for the i-th class. In the cosine feature space, the class probability of i-th class is computed as:
\begin{align}
    p_{\text{cls}}^{\text{cos}} &= \mathrm{Softmax}\left(d_{\text{cos}}\left(\bm{W}_i, \bm{f}\right)\right) \\
    &= \mathrm{Softmax}\left(\frac{\tau \bm{W}_i \bm{f}}{\|\bm{W}_i\|\|\bm{f}\|}\right)
\end{align}
where class probabilities depend entirely on the angular alignment between ROI features and class prototypes, making objectness recognition strongly class-dependent.

In the Euclidean feature space, the class probability of i-th class is defined as:
\begin{equation}
    \begin{aligned}
    p_{\text{cls}}^{\text{euc}} &= \mathrm{Softmax}\left(d_{\text{euc}}\left(\bm{W}_i, \bm{f}\right)\right) \\
    &= \mathrm{Softmax}\left(\bm{W}_i \bm{f}\right) \\
    &= \mathrm{Softmax}\left(\frac{\bm{W}_i \bm{f}}{\|\bm{W}_i\|\|\bm{f}\|} \cdot {\|\bm{W}_i\|\|\bm{f}\|}\right)
\end{aligned}
\end{equation}

And for the orthogonal feature space, the similarity score is formulated as:
\begin{equation}
\begin{aligned}
    p_{\text{cls}}^{\text{OFS}} &=  p_{\text{cls}} \cdot p_{\text{obj}} \\
    &= \mathrm{Softmax}\left(\tau \frac{\bm{W}_i \bm{f}}{\|\bm{W}_i\|\|\bm{f}\|}\right) \cdot \left(-|\bm{W}_{\text{obj}}| \cdot \| \bm{f} \| + b\right)
\end{aligned}
\end{equation}

Compared to cosine feature space, Euclidean feature space and OFS both engage the ROI feature magnitude $\|\bm{f}\|$ into the probability computation. The difference between them is in two folds: (1) In Euclidean space, since the magnitude is not enforced for objectness prediction, detectors can still utilize angular information to determine whether a bounding box is the foreground or the background. (2) Compared to OFS, Euclidean space learns a class-specific term $\|\bm{W}_i\|$ to scale the class probability, which is optimized and overfited to novel class samples. In contrast, OFS learns class-agnostic terms $\|\bm{W}_{\text{obj}}\|$ and $b$ for objectness prediction, which are optimized by all class samples. This design alleviates overfitting to specific classes in FSOD.

\begin{table*}[t]
\centering	
\caption{Comparison results among our method and other FSOD methods on PASCAL VOC. The best single run and the best multiple run results are highlighted in bold.}
\label{tab:voc}
\resizebox{1.0\textwidth}{!}{
\begin{tabular}{lccccccccccccccc}
\toprule
\multicolumn{1}{l|}{\multirow{2}{*}{Method}} & \multicolumn{5}{c|}{Novel Split1}                     & \multicolumn{5}{c|}{Novel Split2}                     & \multicolumn{5}{c}{Novel Split3} \\
\multicolumn{1}{l|}{}                                                                & 1    & 2    & 3    & 5    & \multicolumn{1}{c|}{10}   & 1    & 2    & 3    & 5    & \multicolumn{1}{c|}{10}   & 1    & 2    & 3    & 5    & 10   \\ \midrule
\multicolumn{16}{l}{Single Run:}                                                                                                                                                                                                        \\ \hline
\multicolumn{1}{l|}{TFA w/cos \cite{wang2020frustratingly}}                                                       & 39.8 & 36.1 & 44.7 & 55.7 & \multicolumn{1}{c|}{56.0} & 23.5 & 26.9 & 34.1 & 35.1 & \multicolumn{1}{c|}{39.1} & 30.8 & 34.8 & 42.8 & 49.5 & 49.8 \\
\multicolumn{1}{l|}{FSCE \cite{sun2021fsce}}                                                            & 44.2 & 43.8 & 51.4 & 61.9 & \multicolumn{1}{c|}{63.4} & 27.3 & 29.5 & 43.5 & 44.2 & \multicolumn{1}{c|}{50.2} & 37.2 & 41.9 & 47.5 & 54.6 & 58.5 \\
\multicolumn{1}{l|}{Meta-DETR \cite{zhang2022metadetr}}                                                       & 40.6 & 51.4 & 58.0 & 59.2 & \multicolumn{1}{c|}{63.6} & 37.0 & 36.6 & 43.7 & 49.1 & \multicolumn{1}{c|}{54.6} & 41.6 & 45.9 & 52.7 & 58.9 & 60.6 \\
\multicolumn{1}{l|}{DeFRCN \cite{qiao2021defrcn}}                                                          & 57.0 & 58.6 & 64.3 & 67.8 & \multicolumn{1}{c|}{67.0} & 35.8 & 42.7 & 51.0 & \textbf{54.5} & \multicolumn{1}{c|}{52.9} & 52.5 & 56.6 & 55.8 & 60.7 & 62.5 \\
\multicolumn{1}{l|}{VFA \cite{han2023vfa}}                                                             & 57.7 & 64.6 & 64.7 & 67.2 & \multicolumn{1}{c|}{67.4} & 41.4 & 46.2 & 51.1 & 51.8 & \multicolumn{1}{c|}{51.6} & 48.9 & 54.8 & 56.6 & 59.0 & 58.9 \\
\multicolumn{1}{l|}{FPD \cite{wang2024fpd}}                                                             & 46.5 & 62.3 & 65.4 & 68.2 & \multicolumn{1}{c|}{\textbf{69.3}} & 32.2 & 43.6 & 50.3 & 52.5 & \multicolumn{1}{c|}{\textbf{56.1}} & 43.2 & 53.3 & 56.7 & 62.1 & \textbf{64.1} \\
\multicolumn{1}{l|}{UNP \cite{yan2024unp}}                                                             & 43.7 & 58.3 & 59.8 & 63.7 & \multicolumn{1}{c|}{64.2} & 28.1 & 42.8 & 47.7 & 49.5 & \multicolumn{1}{c|}{50.3} & 38.4 & 49.3 & 53.8 & 57.7 & 58.7 \\
\multicolumn{1}{l|}{FSNA \cite{zhu2024fsna}}                                                            & 43.8 & 47.7 & 50.8 & 57.4 & \multicolumn{1}{c|}{60.3} & 23.9 & 32.3 & 37.9 & 40.2 & \multicolumn{1}{c|}{41.8} & 34.0 & 40.7 & 45.5 & 52.3 & 54.0 \\
\multicolumn{1}{l|}{MPFSOD \cite{ma2024mpfsod}}                                                          & 61.7 & 63.4 & 64.2 & \textbf{68.5} & \multicolumn{1}{c|}{\textbf{69.3}} & 41.9 & 42.2 & 42.6 & 50.6 & \multicolumn{1}{c|}{53.1} & 55.0 & 56.4 & 57.6 & 61.8 & 62.7 \\ \rowcolor{graymy}
\multicolumn{1}{l|}{Ours}                                                   & \textbf{64.3} & \textbf{64.8} & \textbf{66.6} & 68.1 & \multicolumn{1}{c|}{65.9} & \textbf{43.9} & \textbf{46.9} & \textbf{51.7} & 52.2 & \multicolumn{1}{c|}{51.2} & \textbf{60.4} & \textbf{61.9} & \textbf{61.3} & \textbf{64.4} & 63.6 \\ \midrule
\multicolumn{16}{l}{Multiple Runs:}                                                                                                                                                                                                      \\ \hline
\multicolumn{1}{l|}{TFA w/cos \cite{wang2020frustratingly}}    & 25.3 & 36.4 & 42.1 & 47.9 & \multicolumn{1}{c|}{52.8} & 18.3 & 27.5 & 30.9 & 34.1 & \multicolumn{1}{c|}{39.5} & 17.9 & 27.2 & 34.3 & 40.8 & 45.6 \\
\multicolumn{1}{l|}{FSCE \cite{sun2021fsce}}                   & 32.9 & 44.0 & 46.8 & 52.9 & \multicolumn{1}{c|}{59.7} & 23.7 & 30.6 & 38.4 & 43.0 & \multicolumn{1}{c|}{48.5} & 22.6 & 33.4 & 39.5 & 47.3 & 54.0 \\
\multicolumn{1}{l|}{Meta-DETR \cite{zhang2022metadetr}}        & 35.1 & 49.0 & 53.2 & 57.4 & \multicolumn{1}{c|}{62.0} & 27.9 & 32.3 & 38.4 & 43.2 & \multicolumn{1}{c|}{51.8} & 34.9 & 41.8 & 47.1 & 54.1 & 58.2 \\
\multicolumn{1}{l|}{DeFRCN \cite{qiao2021defrcn}}                   & 43.8 & 57.5 & 61.4 & 65.3 & \multicolumn{1}{c|}{67.0} & 31.5 & 40.9 & 45.6 & 50.1 & \multicolumn{1}{c|}{52.9} & 38.2 & 50.9 & 54.1 & 59.2 & 61.9 \\
\multicolumn{1}{l|}{VFA \cite{han2023vfa}}                          & 47.4 & 54.4 & 58.5 & 64.5 & \multicolumn{1}{c|}{66.5} & 33.7 & 38.2 & 43.5 & 48.3 & \multicolumn{1}{c|}{52.4} & 43.8 & 48.9 & 53.3 & 58.1 & 60.0 \\
\multicolumn{1}{l|}{FPD \cite{wang2024fpd}}                    & 37.7 & 51.2 & 59.0 & 64.7 & \multicolumn{1}{c|}{67.8} & 28.7 & 40.0 & 44.3 & 50.2 & \multicolumn{1}{c|}{55.5} & 29.2 & 48.3 & 52.1 & 58.4 & 62.1 \\
\multicolumn{1}{l|}{ECEA \cite{xin2024ecea}}                        & \textbf{55.1} & 60.5 & 62.5 & 63.7 & \multicolumn{1}{c|}{64.0} & \textbf{42.1} & \textbf{47.6} & \textbf{48.4} & \textbf{53.0} & \multicolumn{1}{c|}{\textbf{57.7}} & 39.5 & 47.5 & \textbf{60.7} & \textbf{62.8} & \textbf{66.3} \\
\multicolumn{1}{l|}{OPN \cite{wang2025orthogonal}}             & 53.7 & 58.8 & 60.2 & 65.0 & \multicolumn{1}{c|}{66.9} & 38.7 & 41.8 & 46.4 & 50.1 & \multicolumn{1}{c|}{53.6} & 47.1 & 53.3 & 56.1 & 60.0 & 61.6 \\
\multicolumn{1}{l|}{MMKT \cite{du2025text}}                    & 42.9 & 55.5 & 60.8 & \textbf{66.6} & \multicolumn{1}{c|}{\textbf{68.1}} & 31.1 & 41.7 & 43.8 & 51.3 & \multicolumn{1}{c|}{54.1} & 39.1 & 48.9 & 55.4 & 61.1 & 63.8 \\ \rowcolor{graymy}
\multicolumn{1}{l|}{Ours}                                                   & 52.0 & \textbf{61.1} & \textbf{63.5} & 66.1 & \multicolumn{1}{c|}{66.6} & 36.5 & 44.2 & 47.2 & 49.2 & \multicolumn{1}{c|}{51.0} & \textbf{50.1} & \textbf{58.0} & 59.7 & 62.5 & 64.0 \\ \bottomrule
\end{tabular}}
\end{table*}

\section{Experiments}
\subsection{Datasets}
We evaluate the proposed method on commonly adopted PASCAL VOC \cite{everingham2010voc} and MS COCO \cite{lin2014coco} benchmarks.

\textbf{PASCAL VOC} comprises 20 object classes. Following the standard few-shot evaluation protocol, we adopt three predefined splits to select 5 novel classes (Split1: \textit{bird, bus, cow, motorbike, sofa}, Split2: \textit{aeroplane, bottle, cow, horse, sofa}, Split3: \textit{boat, cat, motorbike, sheep, sofa}). The remaining 15 classes are regarded as base classes. Performance is reported using the novel mean Average Precision at 50\% IoU (nAP50) metric.

\textbf{MS COCO} contains 80 object classes, 20 of which overlap with PASCAL VOC. These overlapping classes are regarded as novel classes, while the remaining 60 classes serve as base classes. We report novel mean Average Precision (nAP) and nAP at 75\% IoU (nAP75) metrics for MS COCO.

\begin{table*}[t]
\caption{Comparison results among our method and other FSOD methods on MS COCO. The best single run and the best multiple run results are highlighted in bold.}
\label{tab:coco}
\resizebox{1.0\textwidth}{!}{
\begin{tabular}{l|cccccccccccc}
\toprule
\multirow{2}{*}{Method} & \multicolumn{2}{c}{1-shot}  & \multicolumn{2}{c}{2-shot}    & \multicolumn{2}{c}{3-shot}    & \multicolumn{2}{c}{5-shot}    & \multicolumn{2}{c}{10-shot}   & \multicolumn{2}{c}{30-shot} \\
                                                               & nAP          & nAP75        & nAP           & nAP75         & nAP           & nAP75         & nAP           & nAP75         & nAP           & nAP75         & nAP          & nAP75        \\ \midrule
\multicolumn{13}{l}{Single run:} \\ \hline
TFA w/cos \cite{wang2020frustratingly}                                                     & 3.4          & 3.8          & 4.6           & 4.8           & 6.6           & 6.5           & 8.3           & 8.0           & 10.0          & 9.3           & 13.7         & 13.4         \\
MPSR \cite{wu2020mpsr}                                                          & 2.3          & 2.3          & 3.5           & 3.4           & 5.2           & 5.1           & 6.7           & 6.4           & 9.8           & 9.7           & 14.1         & 14.2         \\
FADI \cite{cao2021fadi}                                                          & 5.7          & 6.0          & 7.0           & 7.0           & 8.6           & 8.3           & 10.1          & 9.7           & 12.2          & 11.9          & 16.1         & 15.8         \\
DeFRCN \cite{qiao2021defrcn}                                                        & 6.5          & 6.9          & 11.8          & 12.4          & 13.4          & 13.6          & 15.3          & 14.6          & 18.6          & 17.6          & \textbf{22.5}         & 22.3         \\
D\&R \cite{li2023dandr}                                                          & 8.3          & $-$            & 12.7          & $-$             & 14.3          & $-$             & 16.4          & $-$             & 18.7          & $-$             & 21.8         & $-$            \\
Norm-VAE \cite{xu2023normvae}                                                     & 9.5          & 8.8          & 13.7          & 13.7          & 14.3          & 14.2          & 15.9          & 15.3          & 18.7          & 17.8          & \textbf{22.5}         & \textbf{22.4}         \\
FPD \cite{wang2024fpd}  &  $-$ & $-$ & $-$ & $-$ & $-$ & $-$ & $-$  & $-$ & 16.5          & $-$           & 20.1         &       $-$      \\ \rowcolor{graymy}
Ours                                                           & \textbf{10.3}         & \textbf{10.9}         & \textbf{14.9}          & \textbf{15.5}          & \textbf{16.3}          & \textbf{16.5}          & \textbf{18.3}          & \textbf{18.6}          & \textbf{19.8}          & \textbf{20.2}          & 21.8         & 22.0         \\ \midrule
\multicolumn{13}{l}{Multiple runs:} \\ \hline 
TFA w/cos  \cite{wang2020frustratingly}   & 1.9 & 1.7 & 3.9 & 3.6 & 5.1   & 4.8  & 7.0  & 6.5  & 9.1  & 8.8 & 12.1  & 12.0   \\
DeFRCN \cite{qiao2021defrcn}       & 4.8 & 4.4 & 8.5 & 7.8 & 10.7  & 10.3 & 13.5 & 13.0 & 16.7 & 16.7 & 21.0 & 21.4   \\
FPD \cite{wang2024fpd}          &  $-$ & $-$ & $-$ & $-$ & $-$ & $-$ & $-$  & $-$ & 15.9  & $-$ & 19.3 & $-$  \\
ECEA \cite{xin2024ecea}          & 6.1  &  $-$& 10.5 & $-$ & 12.5  & $-$ & 14.9 & $-$ & 18.6 & $-$ & \textbf{22.8} & $-$ \\  \rowcolor{graymy}
Ours          & \textbf{9.3} & \textbf{9.3} & \textbf{13.2} & \textbf{13.4} & \textbf{15.1} & \textbf{15.5} & \textbf{17.2} & \textbf{17.6} & \textbf{19.0} & \textbf{19.7} & 21.0         & \textbf{21.6}         \\ \bottomrule
\end{tabular}}
\end{table*}

\subsection{Implement Details}
We adopt DeFRCN \cite{qiao2021defrcn} as the baseline model, which is built upon the Faster R-CNN detector with a ResNet-101 \cite{he2016resnet} backbone. The loss weight $\alpha$ is set to 0.5. The number of unknown class prototypes $N_u$ is set to 5 unless otherwise specified. During base training, the model is trained for 20,000 iterations on PASCAL VOC and 130,000 iterations on MS COCO. In a minibatch, 8 images from the base set and 8 images from the PB base set are loaded. During fine-tuning of MS COCO, models are finetuned for 800/1,000/1,200/1,500/2,000/2,500 iterations on 1/2/3/5/10/30-shot settings, respectively.

\subsection{Comparison Results}
\label{sec:exp-comparison}
For comparison experiments, we report results from both a single run and the average of 10 runs with different random seeds, consistent with prior work on both PASCAL VOC and MS COCO.

\textbf{PASCAL VOC.} As shown in Table \ref{tab:voc}, the proposed method significantly outperforms the DeFRCN baseline and surpasses recent state-of-the-art approaches, particularly in extreme low-shot scenarios. In high-shot settings of Split 2, our method underperforms compared to DeFRCN. We argue that this degradation comes from the variations in instance sizes. Statistical analysis reveals distinct average instance areas across splits: Split 1 (41,338), Split 2 (32,779), Split 3 (45,028). With \textit{bottle} as the novel class, Split2 novel class instances have the smallest average size. This demonstrates that the proposed method still lacks the ability to recall tiny instances, as optimizing foreground features of varying sizes to the same magnitudes remains challenging. This partly comes from the imbalance of positive/negative samples, where positive samples predominantly have large size.

\textbf{MS COCO.} Comparison results on MS COCO are listed in Table \ref{tab:coco}. Our method achieves superior performance over the baseline DeFRCN across 1/2/3/5/10-shot configurations. Under the multiple run setting, it marginally underperforms DeFRCN with 30 samples per class. These results further demonstrate the robustness of our approach.

\subsection{Ablation Studies}
\label{sec:exp-ablation}

\textbf{Impact of PB base set background design.} As detailed in Section \ref{method:pbbs}, we evaluate three background designs for the PB base set, with results summarized in Table \ref{tab:bg-type}. Simply replacing base image backgrounds with gray pixels (``Gray'') achieves competitive performance. By re-normalizing the gray pixels with foreground pixel mean values (``Normalized Gray''), performance further improves. We attribute this improvement to the introduction of background pixel variation, which prevents the network from relying on pixel averages to distinguish foregrounds from backgrounds. Considering single-color backgrounds are still different from real world scenes, we also attempt to construct the PB base set by pasting foreground instances onto BG-20K background images (``BG-20K''). Surprisingly, this design performs even worse than the gray background. We argue that this occurs because BG-20K adopts a saliency-based method to select background images and still contains a few foreground instances, as filtering all foreground objects using saliency maps remains challenging.

\begin{table}[t]
\centering
\caption{Impact of different background type of PB base set.}
\label{tab:bg-type}
\begin{tabular}{c|ccc}
\toprule
Background      & 1-shot & 2-shot & 3-shot \\ \hline
BG-20K          & 44.7   & 50.7   & 52.4   \\ 
Gray            & 46.3   & 53.4   & 55.6   \\  
Normalized Gray & 47.3   & 55.1   & 56.7   \\ \bottomrule
\end{tabular}
\end{table}

\begin{table}[t]
\centering
\caption{Impact of different feature space.}
\label{tab:fea-space}
\begin{tabular}{c|ccc}
\toprule
Feature Space & 1-shot & 2-shot & 3-shot \\ \hline
Cosine        & 37.7   & 50.5   & 53.6   \\
Euclidean     & 39.3   & 50.6   & 54.9   \\
OFS           & 38.5   & 45.5   & 47.3   \\
UOFS          & 47.3   & 55.1   & 56.7   \\ \bottomrule
\end{tabular}
\end{table}

\begin{table}[!t]
\caption{Silhouette coefficient comparison of orthogonal feature space and uniform orthogonal feature space.}
\centering
\label{tab:silh}
\begin{tabular}{c|cc}
\toprule
\multirow{2}{*}{Feature Space}  & \multicolumn{2}{l}{Silhouette Coefficient} \\
                   & base                 & all                 \\ \hline
OFS                & 0.1991               & 0.1018              \\
UOFS               & 0.2023               & 0.1519              \\ \bottomrule
\end{tabular}
\end{table}

\begin{table}[!t]
\centering
\caption{Impact of the number of unknown class prototype.}
\label{tab:num. learnable bg}
\begin{tabular}{c|ccc}
\toprule
Num. learn bg      & 1-shot & 2-shot & 3-shot \\ \hline
1                  & 44.0   & 51.0   & 53.1   \\ 
3                  & 47.8   & 54.6   & 56.1   \\ 
5                  & 47.3   & 55.1   & 56.7   \\ 
10                 & 45.4   & 53.2   & 55.1   \\ \bottomrule
\end{tabular}
\end{table}

\textbf{Impact of different feature space.} In this paper, we propose the uniform orthogonal feature space (UOFS) for FSOD. We compare it with the commonly adopted cosine/Euclidean feature space and the orthogonal feature space. Results in Table \ref{tab:fea-space} show that the simple OFS is inferior to cosine/Euclidean feature space, while UOFS substantially outperforms OFS. We further analyze clustering quality using the Silhouette Coefficient (SC) in Table \ref{tab:silh}, where higher SC values indicate better cluster separation (smaller intra-class and larger inter-class distances). The results show that UOFS has comparable SC to OFS on base classes but outperforms OFS significantly on all (base+novel) classes. This suggests that UOFS can better generalize to novel classes. The impact of UOFS on base classes is in two folds. On one hand, UOFS prevents unlabeled instances similar to base classes passively cluster to corresponding base classes, creating looser base class distributions. On the other hand, the insertion of learnable unknown class prototypes occupies the original feature space, and the space available for the base class samples becomes smaller, reducing the available space for base class samples and thereby decreasing their intra-class distances. These opposing effects collectively maintain UOFS's base class SC comparable to OFS while improving novel class adaptability. We provide the impact of the number of learnable backgrounds in Table \ref{tab:num. learnable bg}. Results show that 5 learnable backgrounds can achieve the best performance.

\begin{table}[!t]
\centering
\caption{Impact of negative sample placement strategy.}
\label{tab:inner}
\begin{tabular}{c|ccc}
\toprule
place              & 1-shot & 2-shot & 3-shot \\ \hline
inner              & 44.7   & 52.1   & 53.4   \\ 
outer              & 47.3   & 55.1   & 56.7   \\  \bottomrule
\end{tabular}
\end{table}

\begin{figure}[t]
    \centering
    \includegraphics[width=0.95\columnwidth]{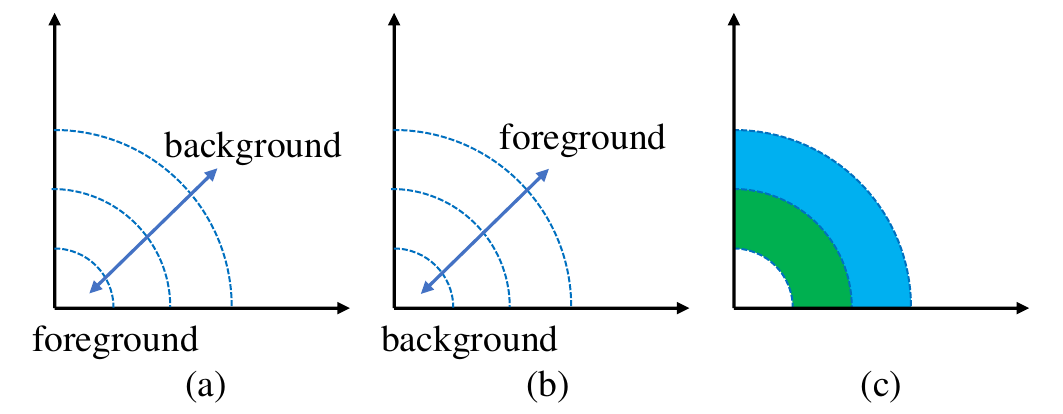}
    \caption{Illustration of the impact of hyperspherical regions.}
    \label{fig: exp-inner}
\end{figure}

\textbf{Impact of negative sample placement strategy.} As shown in Fig. \ref{fig: exp-inner}, in UOFS, we could place negative samples on hyperspheres with larger radius or smaller radius. The comparative results in Table \ref{tab:inner} reveal superior performance when negative samples occupy outer hypersphere regions. This design aligns with the geometric principle: outer hypersphere regions inherently provide larger volumetric capacity to encapsulate diverse feature representations. In the Faster R-CNN detector, positive samples are boxes with IoU larger than 0.5 with ground-truth box while negative samples are boxes with IoU smaller than 0.5. Consequently, negative samples demonstrate greater variability and necessitate allocation within larger hyperspherical regions to accommodate their diverse feature representations. Thus placing negative samples in the outer part is easier to obtain a general feature space. Furthermore, the scarcity of positive samples amplifies overfitting risks when assigned excessive feature space. This is known as the curse of dimensionality. Strategic placement of negative samples in outer regions mitigates this phenomenon through proper spatial resource allocation.

\textbf{Impact of class-agnostic regression.} The proposed framework implements a class-agnostic regression head to compute proposal box offsets. In contrast, the baseline DeFRCN model with the Euclidean feature space employs class-specific regression heads. To evaluate this design choice, we try to adopt class-agnostic regression in the Euclidean space against class-specific regression in UOFS, as detailed in Table \ref{tab:agnreg}. Experimental results indicate that both configurations result in performance degradation. This phenomenon likely stems from task-objective mismatches: in the Euclidean space, where each class occupies a distinct subspace for representation, enforcing class-agnostic regression incorrectly links different class instances. In UOFS, objectness recognition is class-agnostic, adopting class-agnostic regression enables joint optimization.

\begin{table}[!t]
\centering
\caption{Impact of class-agnostic regression.}
\label{tab:agnreg}
\begin{tabular}{c|ccc}
\toprule
 Cls-agn Reg  & 1-shot & 2-shot & 3-shot \\ \hline
 \multicolumn{1}{l}{Eucliden:} \\ \hline
 \ding{56} & 39.3   & 50.6   & 54.9   \\
 \Checkmark   & 36.2   & 47.8   & 51.1   \\ \hline
  \multicolumn{1}{l}{UOFS:} \\ \hline
 \ding{56} & 44.1   & 53.0   & 55.6   \\
 \Checkmark   & 47.3   & 55.1   & 56.7   \\ \bottomrule
\end{tabular}
\end{table}

\begin{table}[!t]
\centering
\caption{Effectiveness of SADA.}
\label{tab:sada}
\begin{tabular}{c|ccc}
\toprule
method   & 1-shot & 2-shot & 3-shot \\ \hline
w/o      & 47.3   & 55.1   & 56.7   \\ 
affine   & 48.0   & 55.4   & 56.8   \\ 
unified  & 45.0   & 53.4   & 54.8   \\
SADA-1   & 47.2   & 54.7   & 56.1   \\
SADA-2   & 48.1   & 56.1   & 58.1   \\
SADA-3   & 50.1   & 58.0   & 59.7   \\ \bottomrule
\end{tabular}
\end{table}

\textbf{Impact of spatial-wise task disentanglement and association.}
In this paper, we propose a spatial-wise task disentanglement and association module to address the task conflicts and associations of \normalsize{\textcircled{\scriptsize{1}}} objectness recognition, \normalsize{\textcircled{\scriptsize{2}}} objectness-agnostic classification and \normalsize{\textcircled{\scriptsize{3}}} box regression. As shown in Table \ref{tab:sada}, we first evaluate spatial attention mechanisms without task disentanglement by employing a shared attention layer across all tasks (``unified''). Results show that the spatial attention itself even brings a performance degradation. In the Faster R-CNN detector, the parameters of backbone and Res5ROIHead are initialized by the ImageNet pretrained ResNet, inserting attention layers between Res4 and Res5 blocks disrupts learned feature hierarchies. With the spatial attention layer, we then test the three designs of SADA. The three tasks are grouped as follow: ``SADA-1'': \normalsize{\textcircled{\scriptsize{1}}}\normalsize{\textcircled{\scriptsize{2}}}, \normalsize{\textcircled{\scriptsize{3}}},  ``SADA-2'': \normalsize{\textcircled{\scriptsize{1}}}, \normalsize{\textcircled{\scriptsize{2}}}, \normalsize{\textcircled{\scriptsize{3}}}, ``SADA-3'': \normalsize{\textcircled{\scriptsize{1}}}\normalsize{\textcircled{\scriptsize{3}}}, \normalsize{\textcircled{\scriptsize{2}}}. ``SADA-1'' is similar to previous works decouple classification and regression, results show that this solution outperforms using the unified spatial layer. ``SADA-2'' generates class-specific features for the three tasks in isolation. ``SADA-3'' achieves the best results because \normalsize{\textcircled{\scriptsize{1}}}\normalsize{\textcircled{\scriptsize{3}}} are class-agnostic while \normalsize{\textcircled{\scriptsize{2}}} is class-specific. Furthermore, replacing ``SADA-3''s spatial attention with the affine transformation in \cite{qiao2021defrcn} yields marginal baseline improvements but underperforms our spatial disentanglement approach. This demonstrates that spatial-wise disentanglement is a more explicit and effective solution.

\begin{figure}[t]
    \centering
    \includegraphics[width=1.0\columnwidth]{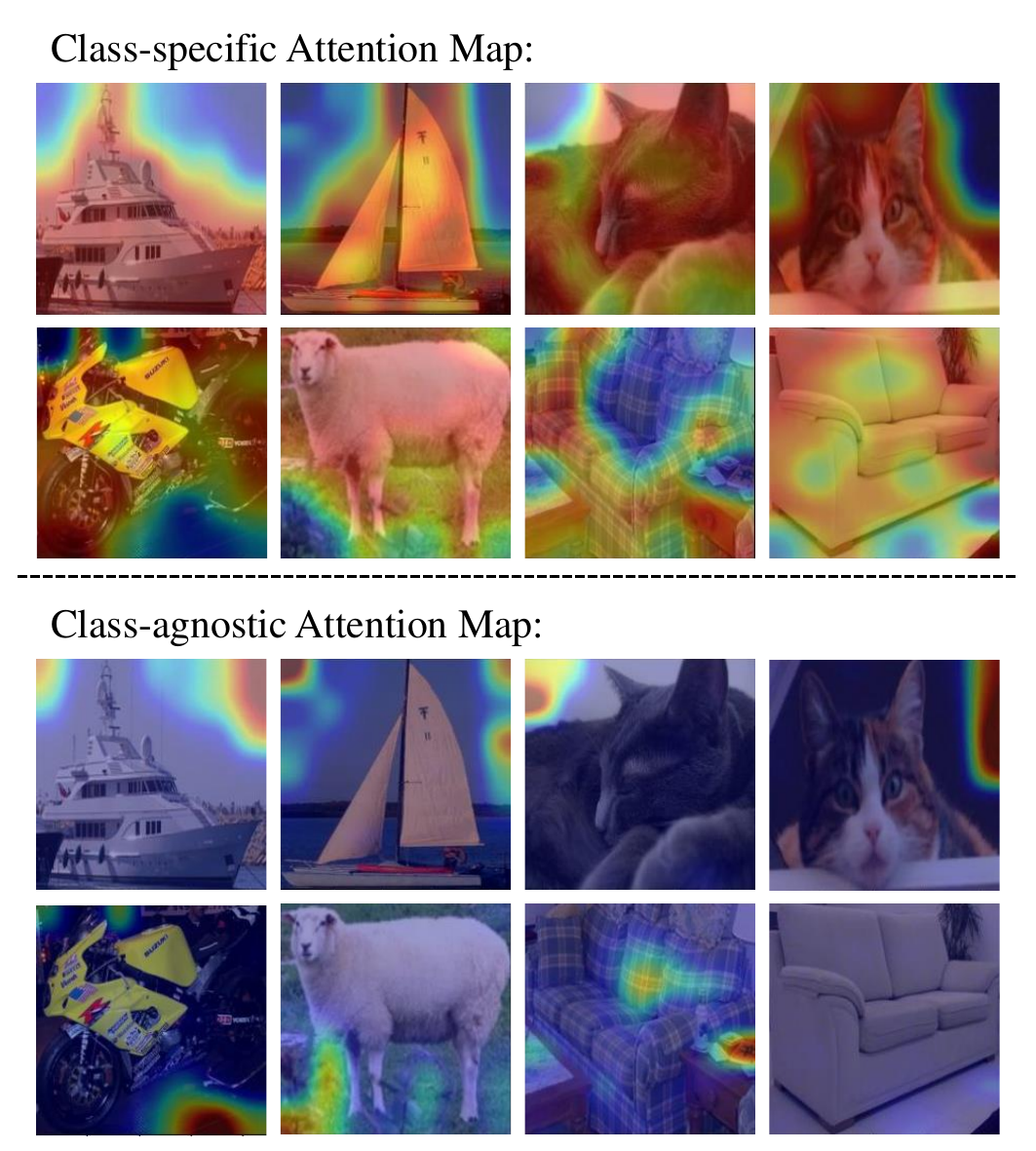}
    \caption{Visualization results of disentangled class-specific attention map for classification and class-agnostic attention map for objectness recognition and box regression.}
    \label{fig: exp-att}
\end{figure}

\subsection{Subjective Results}
\label{sec:exp-subjective}
We present some visualization results of the disentangled attention maps in Fig. \ref{fig: exp-att}. The first two rows show attention maps from the class-specific branch, while the last two rows display those from the class-agnostic branch. By eliminating background intervals, class-specific branch tends to focus on object-centric regions to discriminate the category information within instances. In contrast, class-agnostic branch tends to focus on the whole instances or concentrate on background regions to determine whether a proposal belongs to foreground. Given the flexibility to different tasks, the proposed SADA can better help the network to capture discriminative features.

\section{Conclusion}
\label{sec:conclusion}
In this paper, we address the critical limitations of existing few-shot object detection methods, which suffer from biased objectness criteria and overfitting caused by entangled feature spaces in Faster R-CNN based frameworks. By proposing the Uniform Orthogonal Feature Space (UOFS) optimization framework, we decouple objectness and classification into orthogonal components: magnitude for class-agnostic objectness and angle for class-specific categorization. This disentanglement enables seamless transfer of objectness knowledge from base to novel classes. To mitigate risks arising from unlabeled foreground interference and angular overfitting, we propose Hybrid Background Optimization (HBO), which leverages a pure background base set for unbiased magnitude alignment and incorporates unlabeled instances to enhance angular distribution uniformity. Additionally, our Spatial-wise Attention Disentanglement and Association (SADA) module resolves task conflicts between class-agnostic and class-specific objectives, ensuring robust feature learning. Our method offers a comprehensive solution to feature space entanglement and task misalignment challenges in few-shot object detection.

\bibliographystyle{IEEEtran}
\bibliography{ref}

\begin{thebibliography}{10}
\providecommand{\url}[1]{#1}
\csname url@samestyle\endcsname
\providecommand{\newblock}{\relax}
\providecommand{\bibinfo}[2]{#2}
\providecommand{\BIBentrySTDinterwordspacing}{\spaceskip=0pt\relax}
\providecommand{\BIBentryALTinterwordstretchfactor}{4}
\providecommand{\BIBentryALTinterwordspacing}{\spaceskip=\fontdimen2\font plus
\BIBentryALTinterwordstretchfactor\fontdimen3\font minus \fontdimen4\font\relax}
\providecommand{\BIBforeignlanguage}[2]{{%
\expandafter\ifx\csname l@#1\endcsname\relax
\typeout{** WARNING: IEEEtran.bst: No hyphenation pattern has been}%
\typeout{** loaded for the language `#1'. Using the pattern for}%
\typeout{** the default language instead.}%
\else
\language=\csname l@#1\endcsname
\fi
#2}}
\providecommand{\BIBdecl}{\relax}
\BIBdecl

\bibitem{redmon2017yolo9000}
J.~Redmon and A.~Farhadi, ``Yolo9000: better, faster, stronger,'' in \emph{Proceedings of the IEEE Conference on Computer Vision and Pattern Recognition}, 2017, pp. 7263--7271.

\bibitem{qiu2022crossdet++}
H.~Qiu, H.~Li, Q.~Wu, J.~Cui, Z.~Song, L.~Wang, and M.~Zhang, ``Crossdet++: Growing crossline representation for object detection,'' \emph{IEEE Transactions on Circuits and Systems for Video Technology}, vol.~33, no.~3, pp. 1093--1108, 2022.

\bibitem{girshick2015fast}
R.~B. Girshick, ``Fast {R-CNN},'' in \emph{Proceedings of the IEEE International Conference on Computer Vision}, 2015, pp. 1440--1448.

\bibitem{ren2016faster}
S.~Ren, K.~He, R.~Girshick, and J.~Sun, ``Faster r-cnn: Towards real-time object detection with region proposal networks,'' \emph{IEEE Transactions on Pattern Analysis and Machine Intelligence}, vol.~39, no.~6, pp. 1137--1149, 2016.

\bibitem{cai2018cascade}
Z.~Cai and N.~Vasconcelos, ``Cascade r-cnn: Delving into high quality object detection,'' in \emph{Proceedings of the IEEE Conference on Computer Vision and Pattern Recognition}, 2018, pp. 6154--6162.

\bibitem{wang2020frustratingly}
X.~Wang, T.~Huang, J.~Gonzalez, T.~Darrell, and F.~Yu, ``Frustratingly simple few-shot object detection,'' in \emph{Proceedings of the International Conference on Machine Learning}.\hskip 1em plus 0.5em minus 0.4em\relax PMLR, 2020, pp. 9919--9928.

\bibitem{qiao2021defrcn}
L.~Qiao, Y.~Zhao, Z.~Li, X.~Qiu, J.~Wu, and C.~Zhang, ``Defrcn: Decoupled faster r-cnn for few-shot object detection,'' in \emph{Proceedings of the IEEE International Conference on Computer Vision}, 2021, pp. 8681--8690.

\bibitem{cao2021fadi}
Y.~Cao, J.~Wang, Y.~Jin, T.~Wu, K.~Chen, Z.~Liu, and D.~Lin, ``Few-shot object detection via association and discrimination,'' \emph{Advances in Neural Information Processing Systems}, vol.~34, pp. 16\,570--16\,581, 2021.

\bibitem{kaul2022label}
P.~Kaul, W.~Xie, and A.~Zisserman, ``Label, verify, correct: A simple few shot object detection method,'' in \emph{Proceedings of the IEEE Conference on Computer Vision and Pattern Recognition}, 2022, pp. 14\,237--14\,247.

\bibitem{tang2023semi}
Y.~Tang, Z.~Cao, Y.~Yang, J.~Liu, and J.~Yu, ``Semi-supervised few-shot object detection via adaptive pseudo labeling,'' \emph{IEEE Transactions on Circuits and Systems for Video Technology}, 2023.

\bibitem{zhao2024vlm}
T.~Zhao, H.~Qiu, Y.~Dai, L.~Wang, H.~Mei, F.~Meng, Q.~Wu, and H.~Li, ``Vlm-guided explicit-implicit complementary novel class semantic learning for few-shot object detection,'' \emph{Expert Systems with Applications}, vol. 256, p. 124926, 2024.

\bibitem{xin2024ecea}
Z.~Xin, T.~Wu, S.~Chen, Y.~Zou, L.~Shao, and X.~You, ``Ecea: Extensible co-existing attention for few-shot object detection,'' \emph{IEEE Transactions on Image Processing}, 2024.

\bibitem{sun2024exploring}
Z.~Sun, J.~Li, and Y.~Mu, ``Exploring orthogonality in open world object detection,'' in \emph{Proceedings of the IEEE Conference on Computer Vision and Pattern Recognition}, 2024, pp. 17\,302--17\,312.

\bibitem{chen2020norm}
D.~Chen, S.~Zhang, J.~Yang, and B.~Schiele, ``Norm-aware embedding for efficient person search,'' in \emph{Proceedings of the IEEE Conference on Computer Vision and Pattern Recognition}, 2020, pp. 12\,615--12\,624.

\bibitem{kang2019few}
B.~Kang, Z.~Liu, X.~Wang, F.~Yu, J.~Feng, and T.~Darrell, ``Few-shot object detection via feature reweighting,'' in \emph{Proceedings of the IEEE International Conference on Computer Vision}, 2019, pp. 8420--8429.

\bibitem{li2021beyond}
B.~Li, B.~Yang, C.~Liu, F.~Liu, R.~Ji, and Q.~Ye, ``Beyond max-margin: Class margin equilibrium for few-shot object detection,'' in \emph{Proceedings of the IEEE Conference on Computer Vision and Pattern Recognition}, 2021, pp. 7363--7372.

\bibitem{xin2024few}
Z.~Xin, S.~Chen, T.~Wu, Y.~Shao, W.~Ding, and X.~You, ``Few-shot object detection: Research advances and challenges,'' \emph{Information Fusion}, p. 102307, 2024.

\bibitem{simon2020adaptive}
C.~Simon, P.~Koniusz, R.~Nock, and M.~Harandi, ``Adaptive subspaces for few-shot learning,'' in \emph{Proceedings of the IEEE Conference on Computer Vision and Pattern Recognition}, 2020, pp. 4136--4145.

\bibitem{hersche2022constrained}
M.~Hersche, G.~Karunaratne, G.~Cherubini, L.~Benini, A.~Sebastian, and A.~Rahimi, ``Constrained few-shot class-incremental learning,'' in \emph{Proceedings of the IEEE Conference on Computer Vision and Pattern Recognition}, 2022, pp. 9057--9067.

\bibitem{ranasinghe2021orthogonal}
K.~Ranasinghe, M.~Naseer, M.~Hayat, S.~Khan, and F.~S. Khan, ``Orthogonal projection loss,'' in \emph{Proceedings of the IEEE International Conference on Computer Vision}, 2021, pp. 12\,333--12\,343.

\bibitem{liu2023learning}
S.-A. Liu, Y.~Zhang, Z.~Qiu, H.~Xie, Y.~Zhang, and T.~Yao, ``Learning orthogonal prototypes for generalized few-shot semantic segmentation,'' in \emph{Proceedings of the IEEE Conference on Computer Vision and Pattern Recognition}, 2023, pp. 11\,319--11\,328.

\bibitem{xu2023normvae}
J.~Xu, H.~Le, and D.~Samaras, ``Generating features with increased crop-related diversity for few-shot object detection,'' in \emph{Proceedings of the IEEE Conference on Computer Vision and Pattern Recognition}, 2023, pp. 19\,713--19\,722.

\bibitem{ghiasi2021simple}
G.~Ghiasi, Y.~Cui, A.~Srinivas, R.~Qian, T.-Y. Lin, E.~D. Cubuk, Q.~V. Le, and B.~Zoph, ``Simple copy-paste is a strong data augmentation method for instance segmentation,'' in \emph{Proceedings of the IEEE Conference on Computer Vision and Pattern Recognition}, 2021, pp. 2918--2928.

\bibitem{zhao2023x}
H.~Zhao, D.~Sheng, J.~Bao, D.~Chen, D.~Chen, F.~Wen, L.~Yuan, C.~Liu, W.~Zhou, Q.~Chu \emph{et~al.}, ``X-paste: Revisiting scalable copy-paste for instance segmentation using clip and stablediffusion,'' in \emph{Proceedings of the International Conference on Machine Learning}.\hskip 1em plus 0.5em minus 0.4em\relax PMLR, 2023, pp. 42\,098--42\,109.

\bibitem{dorkenwald2024pin}
M.~Dorkenwald, N.~Barazani, C.~G. Snoek, and Y.~M. Asano, ``Pin: Positional insert unlocks object localisation abilities in vlms,'' in \emph{Proceedings of the IEEE Conference on Computer Vision and Pattern Recognition}, 2024, pp. 13\,548--13\,558.

\bibitem{lin2023effective}
S.~Lin, K.~Wang, X.~Zeng, and R.~Zhao, ``An effective crop-paste pipeline for few-shot object detection,'' in \emph{Proceedings of the IEEE Conference on Computer Vision and Pattern Recognition}, 2023, pp. 4820--4828.

\bibitem{saito2022learning}
K.~Saito, P.~Hu, T.~Darrell, and K.~Saenko, ``Learning to detect every thing in an open world,'' in \emph{Proceedings of the European Conference on Computer Vision}.\hskip 1em plus 0.5em minus 0.4em\relax Springer, 2022, pp. 268--284.

\bibitem{sun2021fsce}
B.~Sun, B.~Li, S.~Cai, Y.~Yuan, and C.~Zhang, ``Fsce: Few-shot object detection via contrastive proposal encoding,'' in \emph{Proceedings of the IEEE Conference on Computer Vision and Pattern Recognition}, 2021, pp. 7352--7362.

\bibitem{wang2025orthogonal}
B.~Wang and D.~Yu, ``Orthogonal progressive network for few-shot object detection,'' \emph{Expert Systems with Applications}, p. 125905, 2025.

\bibitem{ma2023cat}
S.~Ma, Y.~Wang, Y.~Wei, J.~Fan, T.~H. Li, H.~Liu, and F.~Lv, ``Cat: Localization and identification cascade detection transformer for open-world object detection,'' in \emph{Proceedings of the IEEE Conference on Computer Vision and Pattern Recognition}, 2023, pp. 19\,681--19\,690.

\bibitem{ravi2024sam2}
N.~Ravi, V.~Gabeur, Y.-T. Hu, R.~Hu, C.~Ryali, T.~Ma, H.~Khedr, R.~R{\"a}dle, C.~Rolland, L.~Gustafson, E.~Mintun, J.~Pan, K.~V. Alwala, N.~Carion, C.-Y. Wu, R.~Girshick, P.~Doll{\'a}r, and C.~Feichtenhofer, ``Sam 2: Segment anything in images and videos,'' \emph{arXiv preprint arXiv:2408.00714}, 2024.

\bibitem{li2022bridging}
J.~Li, J.~Zhang, S.~J. Maybank, and D.~Tao, ``Bridging composite and real: towards end-to-end deep image matting,'' \emph{International Journal of Computer Vision}, vol. 130, no.~2, pp. 246--266, 2022.

\bibitem{wu2022mfdc}
S.~Wu, W.~Pei, D.~Mei, F.~Chen, J.~Tian, and G.~Lu, ``Multi-faceted distillation of base-novel commonality for few-shot object detection,'' in \emph{Proceedings of the European Conference on Computer Vision}.\hskip 1em plus 0.5em minus 0.4em\relax Springer, 2022, pp. 578--594.

\bibitem{han2023vfa}
J.~Han, Y.~Ren, J.~Ding, K.~Yan, and G.-S. Xia, ``Few-shot object detection via variational feature aggregation,'' in \emph{Proceedings of the AAAI Conference on Artificial Intelligence}, vol.~37, no.~1, 2023, pp. 755--763.

\bibitem{qiu2024mcce}
H.~Qiu, L.~Wang, T.~Zhao, F.~Meng, Q.~Wu, and H.~Li, ``Mcce-rec: Mllm-driven cross-modal contrastive entropy model for zero-shot referring expression comprehension,'' \emph{IEEE Transactions on Circuits and Systems for Video Technology}, 2024.

\bibitem{zhang2022metadetr}
G.~Zhang, Z.~Luo, K.~Cui, S.~Lu, and E.~P. Xing, ``Meta-detr: Image-level few-shot detection with inter-class correlation exploitation,'' \emph{IEEE Transactions on Pattern Analysis and Machine Intelligence}, vol.~45, no.~11, pp. 12\,832--12\,843, 2022.

\bibitem{wang2024fpd}
Z.~Wang, B.~Yang, H.~Yue, and Z.~Ma, ``Fine-grained prototypes distillation for few-shot object detection,'' in \emph{Proceedings of the AAAI Conference on Artificial Intelligence}, vol.~38, no.~6, 2024, pp. 5859--5866.

\bibitem{yan2024unp}
B.~Yan, C.~Lang, G.~Cheng, and J.~Han, ``Understanding negative proposals in generic few-shot object detection,'' \emph{IEEE Transactions on Circuits and Systems for Video Technology}, 2024.

\bibitem{zhu2024fsna}
J.~Zhu, Q.~Wang, X.~Dong, W.~Ruan, H.~Chen, L.~Lei, and G.~Hao, ``Fsna: Few-shot object detection via neighborhood information adaption and all attention,'' \emph{IEEE Transactions on Circuits and Systems for Video Technology}, 2024.

\bibitem{ma2024mpfsod}
J.~Ma and S.~Bai, ``Multi-view part-based few-shot object detection,'' \emph{IEEE Transactions on Neural Networks and Learning Systems}, 2024.

\bibitem{du2025text}
Y.~Du, F.~Liu, L.~Jiao, S.~Li, Z.~Hao, P.~Li, J.~Wang, H.~Wang, and X.~Liu, ``Text generation and multi-modal knowledge transfer for few-shot object detection,'' \emph{Pattern Recognition}, p. 111283, 2025.

\bibitem{everingham2010voc}
M.~Everingham, L.~Van~Gool, C.~K. Williams, J.~Winn, and A.~Zisserman, ``The pascal visual object classes (voc) challenge,'' \emph{International Journal of Computer Vision}, vol.~88, pp. 303--338, 2010.

\bibitem{lin2014coco}
T.-Y. Lin, M.~Maire, S.~Belongie, J.~Hays, P.~Perona, D.~Ramanan, P.~Doll{\'a}r, and C.~L. Zitnick, ``Microsoft coco: Common objects in context,'' in \emph{Proceedings of the European Conference on Computer Vision}.\hskip 1em plus 0.5em minus 0.4em\relax Springer, 2014, pp. 740--755.

\bibitem{wu2020mpsr}
J.~Wu, S.~Liu, D.~Huang, and Y.~Wang, ``Multi-scale positive sample refinement for few-shot object detection,'' in \emph{Proceedings of the European Conference on Computer Vision}, 2020, pp. 456--472.

\bibitem{li2023dandr}
J.~Li, Y.~Zhang, W.~Qiang, L.~Si, C.~Jiao, X.~Hu, C.~Zheng, and F.~Sun, ``Disentangle and remerge: interventional knowledge distillation for few-shot object detection from a conditional causal perspective,'' in \emph{Proceedings of the AAAI Conference on Artificial Intelligence}, vol.~37, no.~1, 2023, pp. 1323--1333.

\bibitem{he2016resnet}
K.~He, X.~Zhang, S.~Ren, and J.~Sun, ``Deep residual learning for image recognition,'' in \emph{Proceedings of the IEEE Conference on Computer Vision and Pattern Recognition}, 2016, pp. 770--778.

\end{thebibliography}

\end{document}